\documentclass[letterpaper, 10pt, conference]{IEEEtran}

\IEEEoverridecommandlockouts
\usepackage{amsmath,amsfonts,amssymb,amsthm}
\usepackage{algorithm,algpseudocode}
\usepackage{array}
\usepackage[caption=false, font=footnotesize]{subfig}
\usepackage{textcomp}
\usepackage{stfloats}
\usepackage{url}
\usepackage{verbatim}
\usepackage{graphicx}
\usepackage{balance}
\usepackage[percent]{overpic}

\usepackage{siunitx}

\usepackage{booktabs}
\usepackage{multirow}

\theoremstyle{definition}
\newtheorem{definition}{Definition}
\theoremstyle{plain}

\usepackage{flushend}

\begin{document}
\title{\LARGE\bf Rocket Landing Control with Random Annealing Jump Start Reinforcement Learning}

\author{%
Yuxuan~Jiang\textsuperscript{2},
Yujie~Yang\textsuperscript{2},
Zhiqian~Lan\textsuperscript{2},
Guojian~Zhan\textsuperscript{2},
Shengbo~Eben~Li\textsuperscript{*1,2},\\
Qi~Sun\textsuperscript{2},
Jian~Ma\textsuperscript{3},
Tianwen~Yu\textsuperscript{3},
Changwu~Zhang\textsuperscript{3}%
\thanks{This study is supported by Tsinghua University Initiative Scientific Research Program, and NSF China with U20A20334 and 52072213.}%
\thanks{\textsuperscript{1}College of Artificial Intelligence, Tsinghua University, Beijing, 100084, China. \textsuperscript{2}School of Vehicle and Mobility, Tsinghua University, Beijing, 100084, China. \textsuperscript{3}LandSpace Technology Corporation, Beijing, 100176, China.
\textsuperscript{*}All correspondences should be sent to S.~E.~Li with e-mail: lisb04@gmail.com}
}

\maketitle
\thispagestyle{empty}
\pagestyle{empty}

\begin{abstract}
Rocket recycling is a crucial pursuit in aerospace technology, aimed at reducing costs and environmental impact in space exploration. The primary focus centers on rocket landing control, involving the guidance of a nonlinear underactuated rocket with limited fuel in real-time. This challenging task prompts the application of reinforcement learning (RL), yet goal-oriented nature of the problem poses difficulties for standard RL algorithms due to the absence of intermediate reward signals. This paper, for the first time, significantly elevates the success rate of rocket landing control from 8\% with a baseline controller to 97\% on a high-fidelity rocket model using RL. Our approach, called Random Annealing Jump Start (RAJS), is tailored for real-world goal-oriented problems by leveraging prior feedback controllers as guide policy to facilitate environmental exploration and policy learning in RL. In each episode, the guide policy navigates the environment for the guide horizon, followed by the exploration policy taking charge to complete remaining steps. This jump-start strategy prunes exploration space, rendering the problem more tractable to RL algorithms. The guide horizon is sampled from a uniform distribution, with its upper bound annealing to zero based on performance metrics, mitigating distribution shift and mismatch issues in existing methods. Additional enhancements, including cascading jump start, refined reward and terminal condition, and action smoothness regulation, further improve policy performance and practical applicability. The proposed method is validated through extensive evaluation and Hardware-in-the-Loop testing, affirming the effectiveness, real-time feasibility, and smoothness of the proposed controller.
\end{abstract}

\section{Introduction}
Rocket recycling is a pivotal pursuit in the field of aerospace technology, driven by its potential to significantly reduce costs and mitigate the environmental impact of space exploration. Central to this endeavor is the challenge of rocket landing control, a task that involves guiding a nonlinear underactuated rocket plant with limited fuel to land in real-time. The controller must overcome large random disturbances, and satisfy strict terminal state requirement at landing. Conventional control strategies, such as PID controllers, require tedious tuning to meet the accuracy prerequisites, while optimization methods face difficulties in meeting the real-time constraints of this task. Reinforcement learning (RL), employing offline training and online implementation (OTOI) model, presents a viable solution for addressing these complex challenges.

RL offers a powerful paradigm to iteratively optimize policies for control problems by maximizing the expected cumulative rewards. It has shown remarkable success in various domains, including video games \cite{Kaiser2020}, board games \cite{MuZero}, robotics \cite{Andrychowicz2020}, and autonomous driving \cite{Kiran2021}. In these problems, well-crafted dense reward signals are critical for RL agents to progressively improve their policies. However, rocket landing poses a unique challenge for RL. It is a \emph{goal-oriented} problem, requiring the agent to reach a specific goal set in the state space, with intermediate reward signals not readily available. Standard RL algorithms, such as PPO \cite{PPO} , SAC \cite{pmlr-v80-haarnoja18b} and DSAC \cite{duan2021distributional}, fail in such scenarios due to their reliance on extensive exploration; the likelihood of randomly reaching the goal diminishes exponentially over time.

\begin{figure}[t]
    \centering
    \includegraphics[width=0.80\linewidth,trim=1em 1em 1em 2em]{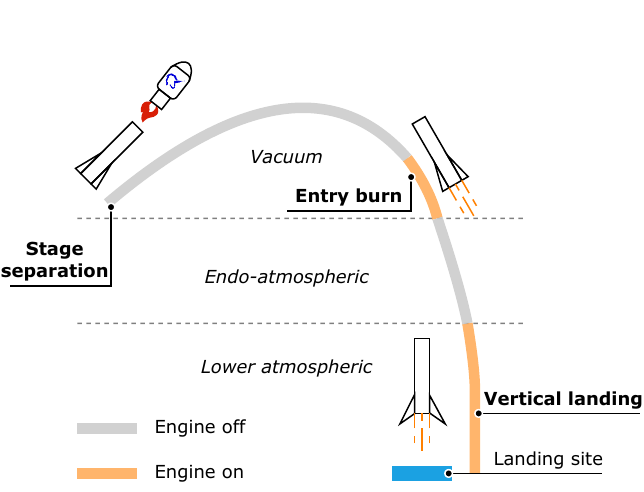}
    \caption{Brief task demonstration of rocket landing control}
    \label{fig:rocket}
    \vspace{-1.8em}
\end{figure}

General goal-oriented tasks without any prior knowledge present intrinsic difficulties. Several algorithm categories, such as intrinsic reward methods \cite{jaderberg2016reinforcement,pathak2017curiosity} and efficient exploration methods \cite{zanette2020learning, xie2021policy} have been developed to optimize policies in such contexts. However, they still face significant challenges in scenarios with hard-to-reach goals and large continuous action spaces, requiring an exponential increase in sample complexity. In many cases, some prior knowledge of the target environment does exist, and leveraging this knowledge can accelerate learning and enhance performance. Reward shaping is a common method for integrating such knowledge. Although manually designed rewards are adaptable and straightforward to implement, they necessitate substantial effort to balance multiple factors. In OpenAI Five \cite{berner2019dota}, for instance, over twenty shaped reward components are carefully weighted to achieve professional-level performance. Another approach, prevalent in robotic control tasks, involves using a reference trajectory with a quadratic tracking reward. For rocket landing control, designing feasible trajectories demands expert knowledge \cite{anglim2016minimum}, and a single fixed trajectory becomes impractical under variable initial conditions and environmental disturbances. Alternatively, some methods utilize more specific prior knowledge, like offline datasets or prior policies, to aid exploration and learning. This class of methods typically target initialization, either initializing the policy before RL training \cite{nair2020awac, lu2022aw} or initializing the state before each episode \cite{nair2018overcoming, agarwal2020pc}. One simple and effective approach is jump start RL (JSRL) \cite{uchendu2022jump}, which initialize each episode by applying a guide policy over the guide horizon before transitioning to the exploration RL policy for the remainder of the task. The effectiveness of the jump start technique hinges on the design of the guide horizon, with two variants, JSRL-Curriculum and JSRL-Random, proposed in JSRL paper.

In this paper, we build on the jump start framework, and introduce the Random Annealing Jump Start (RAJS) approach. RAJS is particularly suitable for real-world control problems like rocket landing, which often involve some form of prior feedback controllers based on PID or other classical control methods. For each episode, RAJS samples the guide horizon uniformly between zero and an upper bound, and anneals the upper bound to zero during training. This minimizes state distribution shifts, addressing the limitations of JSRL-Curriculum and broadening the applicability of RL algorithms from offline, off-policy to on-policy variants. Additionally, the final initial state distribution aligns with the underlying environment, countering the distribution mismatch issue seen in JSRL-Random. The stability afforded by RAJS simplifies the annealing schedule, allowing for either automatic scheduling based on training metric feedback or manual adjustment using a clamped ramp function. We also integrate several generally applicable practical techniques, including cascading jump start, refined reward and terminal condition design, and action smoothness regulation. These enhancements enable us to effectively tackle the rocket landing control problem, elevating the success rate from 8\% with the baseline controller to an impressive 97\%. Extensive evaluation reveals that the control maneuvers are both smooth and interpretable, and Hardware-in-the-Loop testing confirms the real-time feasibility of the control signals produced by our controller.

\section{Preliminary}
\subsection{Problem Statement}
Illustrated in Fig.~\ref{fig:rocket}, our study focuses on the critical phase of vertical landing within the lower atmospheric layer, which presents the highest demand for control precision in rocket recycling missions. In the context of controller operations, the fundamental objective of the rocket landing task is to guide the rocket from diverse initial states to land on the ground, while satisfying various terminal state constraints. The plant is a high-fidelity rocket model built in Simulink from LandSpace, containing calibrated subsystems for inertia, engine, actuator, aerodynamics and disturbance. The coordinate system used for the rocket landing problem has its origin on the ground. The $y$-axis points vertically upward, the $x$-axis points toward the north, and the $z$-axis points toward the east.

The task consists of two stages. On the first stage, all three engines of the rocket are available for deceleration and pose control. The task switches to the second stage when the vertical velocity $v_y$ reaches a threshold of $v_{sw} = \qty{-60}{m/s}$, where the internal low-level control mechanism closes two of three engines, and continue the mission until landing. This internal switch does not affect the controller-rocket interaction directly, but would be reflected in the model dynamics. 

The key states of the rocket plant include position $x$, $y$, $z$, velocity $v_x$, $v_y$, $v_z$, angular position $\phi$, $\psi$, $\gamma$, angular velocity $\dot\phi$, $\dot\psi$, $\dot\gamma$, and total mass $m$, which dynamically reduces as fuel is consumed. Each simulation run allows these state variables to initialize within specified ranges, as shown in Table \ref{tab:initial-state}. The constraints for the terminal state are defined to ensure a precise landing within a narrow vicinity of the target, along with parameters critical for maintaining landing stability, also delineated in Table \ref{tab:initial-state}. The state observation includes 12 kinematic variables (encompassing position, velocity, angular position, and angular velocity) and the axial load. It is important to note that certain plant states, such as those related to engine response, remain hidden from the controller. The control inputs are modeled as three engine attitude signals and one engine thrust signal, normalized within \numrange{-1}{1}. The simulation terminates upon the occurrence of one of the following events: either a successful or failed landing, fuel exhaustion, and vertical speed reversal.

\begin{table}[t]
    \centering
    \caption{Main states, initial values and terminal constraints}
    \label{tab:initial-state}
    \begin{tabular}{ccr@{~$\pm$~}lr@{~$\pm$~}l}
        \toprule
        State & Unit & \multicolumn{2}{c}{Initial value \& range} & \multicolumn{2}{c}{Terminal value \& range} \\
        \midrule
        $x$             & m                     & 50                    & 500   & 0 &5 \\
        $y$             & m                     & 2000                  & 10    & \multicolumn{2}{c}{0} \\
        $z$             & m                     & 0                     & 500   & 0&5 \\
        $v_x$           & m/s                   & $-10$                 & 50    & 0&1 \\
        $v_y$           & m/s                   & $-300$                & 50    & \multicolumn{2}{c}{\hspace{-0.9em}$-1 \sim 0$} \\
        $v_z$           & m/s                   & 0                     & 50    & 0&1 \\
        $\phi$          & $\unit{\degree}$      & 90                    & 0.5   & \hspace{2.75em} 90&3 \\
        $\psi$          & $\unit{\degree}$      & 0                     & 0.5   & 0&3 \\
        $\gamma$        & $\unit{\degree}$      & 0                     & 0.5   & \multicolumn{2}{c}{-} \\
        $\dot\phi$      & $\unit{\degree/s}$    & 0                     & 0.1   & 0&1.5 \\
        $\dot\psi$      & $\unit{\degree/s}$    & 0                     & 0.1   & 0&1.5 \\
        $\dot\gamma$    & $\unit{\degree/s}$    & 0                     & 0.1   & \multicolumn{2}{c}{-} \\
        $m$             & kg                    & \hspace{1.05em}50000  & 500   & \multicolumn{2}{c}{$\geq 40000$} \\
        \bottomrule
    \end{tabular}
    \vspace{-1.8em}
\end{table}

Additional complexity is introduced in the form of wind disturbance, characterized by the wind speed that is uniformly sampled within a range of \qtyrange{0}{15}{m/s}, and the wind direction from any direction within the horizontal plane. Crucially, these wind parameters remain unobservable to the controller, necessitating a control algorithm that possesses robustness to such external perturbations.

A baseline controller is integrated within the plant for establishing a closed-loop simulation environment. This controller, based on several reference trajectories and PID control mechanisms, demonstrates the capability to maintain pose stability and manage the vertical descent under normal conditions. However, its limited adaptability in the face of disturbances results in frequent constraint violations and a modest overall success rate of 8\%. Hence, although the baseline controller offers valuable reference for policy learning, the algorithm must refrain from mere imitation and instead employ it strategically.

\subsection{Goal-oriented Markov decision process}
Reinforcement Learning (RL) is based on the concept of a Markov Decision Process (MDP), where the optimal action solely depends on the current state \cite{li2023rlbook}. To elaborate further, at each time step denoted as $t$, the agent selects an action $a_t \in \mathcal{A}$ in response to the current state $s_t \in \mathcal{S}$ and the policy $\pi: \mathcal{S} \rightarrow \mathcal{A}$. Subsequently, the environment transits to the next state according to its dynamics, represented as $s_{t+1} = f(s_t,a_t)$, and provides a scalar reward signal denoted as $r_t$. The value function $v^{\pi}: \mathcal{S} \rightarrow \mathbb{R}$ is defined as the expected cumulative sum of rewards generated by following policy $\pi$ when initiated from an initial state $s$.

In the context of rocket landing control, we can abstractly model it as a goal-oriented MDP. The agent is required to reach the goal set $\mathcal{S}_{\text{goal}}$ from the initial set $\mathcal{S}_\text{init}$ while maximizing discounted return defined as:
\begin{equation*}
    \max_{\pi} \mathbb{E}_{\tau\sim\pi}\left[\sum_{t=0}^{T} \gamma^t r(s_t)\right],
\end{equation*}
where $\gamma$ is the discount factor, and the goal-oriented reward function is expressed as:
\begin{equation*}
    r(s_t) = \begin{cases}
        1 & s_t \in \mathcal{S}_{\text{goal}}\\
        0 & s_t \notin \mathcal{S}_{\text{goal}}
    \end{cases}.
\end{equation*}
An episode terminates when state $s_t$ enters $\mathcal{S}_{\text{term}} = \mathcal{S}_{\text{goal}} \cup \mathcal{S}_{\text{fail}}$, where $\mathcal{S}_{\text{fail}}$ represents a collection of states disjoint with $\mathcal{S}_{\text{goal}}$ where continuation of the simulation is infeasible.

\subsection{Proximal policy optimization}
Proximal policy optimization (PPO) \cite{PPO} is an on-policy RL algorithm rooted in the policy gradient method, featuring straightforward implementation, parallel sampling, and stability during training. PPO finds extensive application in addressing complex challenges, such as high-dimensional continuous control tasks, the development of advanced AI for professional-level gaming, and the recent advancements in reinforcement learning through human feedback for refining large language models. The core of PPO lies in its objective function, as expressed by:
\begin{equation}
\label{eqn:ppo-loss}
\begin{aligned}
    J_\pi(\theta) = \frac{1}{|\mathcal{D}|} \sum_{x,u\in\mathcal{D}} \min\left(\rho(\theta) A^{\pi_{\text{old}}}(x,u), \right.\\
    \left.\text{clip}\left(\rho(\theta), 1-\epsilon, 1+\epsilon \right) A^{\pi_{\text{old}}}(x,u)\right).
\end{aligned}
\end{equation}
Here, $\theta$ denotes the parameter of the policy network, $\rho(\theta) = \frac{\pi_\theta(u|x)}{\pi_{\text{old}}(u|x)}$ represents the importance sampling factor, and $A^{\pi_{\text{old}}}$ is the advantage function computed using generalized advantage estimation \cite{schulman2015high}. PPO also incorporates the learning of a state value network to estimate $v_\pi$, which serves as a baseline in generalized advantage estimation, contributing to variance reduction.
While PPO excels in exploration capabilities, it encounters significant challenges when addressing goal-oriented environments independently, primarily due to the exponential sample complexity inherent in such scenarios. To overcome this obstacle effectively, it is necessary to combine PPO with our proposed RAJS method, thus facilitating the efficient attainment of satisfactory solutions.

\section{Method}

\subsection{Jump start framework}
The jump start framework comprises a fixed guide policy $\pi_g(a|s)$ and a learnable exploration policy $\pi_e(a|s)$. In each episode, the guide policy first navigates the environment for the guide horizon $H$ steps, after which the exploration policy $\pi_e$ takes over to complete the remaining steps. Selecting $H$ close to the full task horizon reduces the exploration space of $\pi_e$, enhancing the likelihood of reaching the goal.

The effectiveness of the jump start framework hinges on the design of the guide horizon $H$. In essence, jump start modifies the initial state distribution $d_\text{init}$ of the original MDP to an easier $\Tilde{d}_\text{init}$ produced by the guide policy. $\Tilde{d}_\text{init}$ depends solely on the guide horizon $H$, which can be a constant or a random variable. The gap between $\Tilde{d}_\text{init}$ during training and $d_\text{init}$ during practical evaluation can negatively impact policy performance, necessitating careful design of $H$. The distribution mismatch coefficient (Definition \ref{def:dmc}) is a useful metric to quantify such gap.
\begin{definition}[Distribution mismatch coefficient \cite{agarwal2021theory}]\label{def:dmc}
The distribution mismatch coefficient $D_\infty$:
\begin{equation*}
    D_\infty = \left\lVert \frac{d_\rho^\pi}{\mu}\right\rVert_\infty
\end{equation*}
quantifies the effect of policy gradient learning under a initial state distribution $\mu$ possibly deviated from the initial state distribution of interest $\rho$. $d_\rho^\pi$ refer to the discounted stationary state distribution under policy $\pi$:
\begin{equation*}
    d_\rho^\pi = \mathbb{E}_{s_0\sim \rho}\left[(1-\gamma)\sum_{t=0}^\infty{\gamma^t \text{Pr}^\pi(s_t = s | s_0)}\right].
\end{equation*}
\end{definition}

The JSRL paper introduces two variants: JSRL-Curriculum and JSRL-Random. JSRL-Curriculum dynamically adjusts $H$ as a function of training progress using a pre-designed curriculum. In practical terms, a step function is employed:
\begin{equation*}
    H_k = \left(1 - \frac{k}{n}\right) \bar{H}, \quad k = 0, 1, \ldots, n,
\end{equation*}
where $\bar{H}$ is the initial guide horizon, $n$ is the number of curriculum stages, and $k$ is incremented based on the moving average of evaluated performance. The modified initial state distribution $\Tilde{d}_{\text{init},k}$ associated with $H_k$ ensures that $\Tilde{d}_{\text{init},n}$ is equivalent to $d_\text{init}$, benefiting evaluation performance. However, the transition between stages introduces a notable flaw of distribution shift, adding unseen states to the space and changing the initial state distribution at each stage switch. JSRL paper focuses on transitioning from offline RL to online RL, employing implicit Q Learning (IQL) \cite{kostrikov2022offline} as the optimization algorithm in both stages. While IQL mitigates the distribution shift to some extent through its replay buffer, on-policy algorithms like PPO face significant challenges in policy updates due to this shift. Particularly in tasks such as rocket landing control, where initial state distributions $\Tilde{d}_{\text{init}}$ for different guide horizons $H$ can be completely disjoint, this would disrupt policy learning, rendering the current policy unlearned.

On the other hand, JSRL-Random samples the guide horizon $H$ from a uniform distribution $U(0, \bar{H})$, where $\bar{H}$ is the horizon upper bound. Unlike JSRL-Curriculum, the initial state distribution induced by the random variable $H$ remains fixed throughout training, essentially creating an alternative stationary MDP with an adjusted initial state distribution. This approach enhances stability during training compared to JSRL-Curriculum but introduces a different challenge of distribution mismatch. Although it is well-established that the optimal policy $\pi^*$ for an MDP is independent of specific $d_{\text{init}}$ under mild conditions, different $d_{\text{init}}$ do affect the iteration complexity, particularly when function approximation is involved. As proven by \cite{agarwal2021theory}, the suboptimality $V^*(s) - V^\pi(s)$ of policy gradient after $T$ iterations is positively correlated to $D_\infty \epsilon_\text{approx}$. Here, $\epsilon_\text{approx}$ is the approximation error, representing the minimal possible error for the given parametric function to fit the target distribution. This correlation implies that, for a fixed positive $\epsilon_\text{approx}$ of a neural network function approximation, a larger distribution mismatch coefficient $D_\infty$ contributes to slower convergence. In the context of solving goal-oriented problems with jump start, $\rho$ corresponds to $d_\text{init}$, and $\mu$ corresponds to $\Tilde{d}_\text{init}$ of the chosen jump start schedule. Within the support of $d_\text{init}$, JSRL-Random's $\Tilde{d}_\text{init}$ is much smaller than $d_\text{init}$, leading to a larger $D_\infty$. This explains the experimental observation that JSRL-Random's policy performance on the evaluation distribution is significantly inferior to optimality.

\subsection{Random annealing jump start}
Building upon the insights gained from the analysis presented earlier, and considering the effectiveness of the jump start framework in challenging goal-oriented environments when used with on-policy algorithms, we introduce Random Annealing Jump Start (RAJS). This approach addresses the limitations of distribution shift and distribution mismatch observed in prior works. RAJS achieves this by sampling the guide horizon from a uniform distribution, similar to JSRL-Random. However, a key distinction lies in initializing the upper bound of the uniform distribution with a large value and gradually annealing it to 0 during training, as expressed by the equation:
\begin{equation}
    H \sim U(0, \bar{H}\beta(\cdot)),
\end{equation}
where $\beta(\cdot)$ denotes an annealing factor transitioning from 1 to 0. RAJS effectively mitigates distribution mismatch, as the exploration policy directly engages with the underlying goal-oriented environment after the annealing of $\bar{H}$ to 0. Furthermore, RAJS significantly reduces distribution shift compared to JSRL-Curriculum. In the context of $D_\infty$, assuming $\mu$ and $\rho$ represent $\Tilde{d}_\text{init}$ before and after a distribution shift, RAJS exhibits a substantial overlap between $\mu$ and $\rho$, resulting in a much smaller $D_\infty$ in comparison to JSRL-Random.

Due to the minimal distribution shift during training, the tuning of $\beta(\cdot)$ can be simplified. We propose a schedule based on the moving average of training metrics. For goal-oriented tasks, the proportion of episodes terminating in the goal set (or success rate), denoted as $P_\text{goal}$, serves as a suitable metric. The update rule for $\beta(P_\text{goal})$ at the end of each training iteration is as follows:
\begin{equation}
    \beta \gets \max(\beta - \alpha\mathbb{I}(P_\text{goal} \geq P_\text{thresh}), 0),
\end{equation}
where $P_\text{thresh}$ a tunable performance threshold, $\mathbb{I}(\cdot)$ is the indicator function, and $\alpha$ is the update step size. Additionally, due to improved training stability, designing a ramp schedule $\beta(N)$ manually becomes trivial, with $N$ representing the total number of environment interactions. The start and end steps of the ramp can be determined by solving the task once with $\beta\equiv1$ and observing the success rate training curve.

RAJS relaxes JSRL's guide policy requirements, extending its applicability to on-policy RL algorithms, as outlined in Algorithm \ref{alg:RAJS}. In this paper, our focus is on PPO discussed in the preliminary section, leveraging its general applicability to address the complex challenge of rocket landing control.

\begin{algorithm}
\small
\caption{Random annealing jump start w/ on-policy RL}
\label{alg:RAJS}
\begin{algorithmic}[1]
\State \textbf{Input:} guide policy $\pi_g$, maximum guide horizon $\bar{H}$, metric threshold $P_\text{thresh}$, annealing step size $\alpha$, training batch size $B$. 
\State Initialize exploration-policy $\pi_e$, and other required function approximation, e.g., state value function $V$. Initialize annealing factor $\beta \gets 1$, and moving mean metric $P \gets 0$.
\Procedure{Rollout}{$\pi_g$, $\pi_e$, $\bar{H}$, $\mathcal{D}$, $P$}
  \State Sample initial state $s_0 \sim d_\text{init}$, guide horizon $H \sim U(0, \bar{H})$.
  \State Rollout $\lfloor H \rfloor$ steps with guide policy $\pi_g$.
  \State Rollout until termination with exploration policy $\pi_e$, logging trajectory $\{(s, a, r), \ldots\}$ to $\mathcal{D}$.
  \State Update $P$ with metric of current episode.
\EndProcedure
\Repeat
    \State Initialize trajectory dataset $\mathcal{D} = \{\}$.
    \State Sample with \Call{Rollout}{$\pi_g$, $\pi_e$, $\bar{H}\beta$, $\mathcal{D}$, $P$}, until $|\mathcal{D}|\geq B$.
    \State $\pi_e, V \gets \Call{TrainPolicy}{\pi_e, V, \mathcal{D}}$.
    \State $\beta \gets \max(\beta - \alpha\mathbb{I}(P \geq P_\text{thresh}), 0)$.
\Until $\beta = 0$ and convergence
\end{algorithmic}
\end{algorithm}

\subsection{Practical techniques}
Several practical techniques are introduced to further enhance the resolution of goal-oriented tasks. While these techniques were initially developed in the context of rocket landing control, their underlying principles are broadly applicable to a diverse range of tasks.

\subsubsection{Cascading jump start}
In rocket landing control, the baseline controller, which serves as the guide policy, fails in 92\% of cases. Consequently, the exploration policy may face a dilemma of hard exploration or unrecoverable failure, depending on the shorter or longer guide horizon length. Although RAJS significantly reduces the exploration space for RL agents, the initial training phase remains challenging to explore. With respect to the PPO agent, it may suffer from premature entropy collapse, necessitating careful selection of the entropy regulation coefficient to maintain agent exploration. This complicates further performance tuning, as additional difficulty imposed on the agent (e.g., action smoothness requirement) can impede policy convergence.

In this scenario, an effective technique involves incorporating cascading jump start, where an agent is trained under the vanilla setting and subsequently used as the new guide policy $\pi_g'$ in experiments with more demanding settings. Experiments demonstrate that this technique facilitates simplified exploration at the outset of training, with no significant impact on the final performance after convergence.

\subsubsection{Reward design}
The technical relevance of the outcome of a goal-oriented problem lies solely in the goal set, specifically the satisfaction of terminal state constraints associated with the set. As formulated in the preliminary section, the precisely equivalent terminal reward signal would be a binary signal indicating constraint satisfaction. However, the absence of a smooth reward gradient causes the policy's exploration to be purely driven by ``luck", leading to a deterioration in training efficiency and performance.

In an effort to relax the reward, several rules must be followed to ensure that the new objective stays close to the original definition:
\begin{enumerate}
\item All intermediate steps should have zero reward. Non-zero intermediate rewards are prone to unexpected policy exploitation behavior and tend to significantly deviate the task from the original definition.
\item Terminal reward should be non-negative. Easily obtained negative rewards would drive the policy to extend intermediate steps and avoid reaching terminal states due to $\gamma$-discounting.
\end{enumerate}
Adhering to these requirements, we can provide smooth rewards in $\mathcal{S}_{\text{prox}}$, where $\mathcal{S}_{\text{goal}} \subset \mathcal{S}_{\text{prox}} \subseteq \mathcal{S}_{\text{term}}$, to guide trajectories that terminate in the proximity of the goal set in the correct optimization direction. In rocket landing control, $\mathcal{S}_{\text{prox}}$ can be chosen as all landing states $\{s \;|\; y = 0\}$, while other conditions of failure, including fuel exhaustion and vertical speed reversal, still receives zero terminal reward. In $\mathcal{S}_{\text{prox}}$, we propose a logarithmic function:
\begin{equation}
    \label{eqn:log-reward}
    r_{\text{prox}} = \max(b-\log(1 + p \max e), 0),
\end{equation}
where $p$ and $b$ are the scale and bias terms controlling the effective range of terminal states to receive a positive reward. Each element of $e$ corresponds to the normalized absolute terminal error of a constrained state:
\begin{equation*}
    e = \left|\frac{s_T - s_\text{target}}{s_\text{range}}\right|.
\end{equation*}
Compared to the typical quadratic reward form, this ensures a non-zero gradient even if the terminal state is far from the target, as well as a larger gradient close to the target to facilitate constraint satisfaction among noisy samples.

\subsubsection{Terminal condition}
In addition to the primary terminal condition, the application of early termination based on heuristics proves beneficial for credit assignment in the absence of intermediate rewards, leading to an acceleration in policy learning. In the rocket landing task, learning vertical speed control from scratch is time-consuming due to a long control horizon, difficulties in credit assignment, the coupling of $y$ and $v_y$, and the tight bound of $v_y$ when $y$ reaches 0. To address this, kinematics rules are employed to provide coarse information about feasibility. An additional early termination condition is derived for situations where the task is inevitably bound to fail:
\begin{equation}
    y_{\min} = \begin{cases}
        \frac{v_y^2-v_{sw}^2}{2a_{\max,1}} + \frac{v_{sw}^2}{2a_{\max,2}} & v_y \leq v_{sw}, \\
        \frac{v_y^2}{2a_{\max,2}}, & v_{sw} < v_y < 0,
    \end{cases}
\end{equation}
where $a_{\max,1}$ and $a_{\max,2}$ represent the approximate values of maximum deceleration in two control stages. The episode is terminated with zero terminal reward once $y \leq y_{\min}$, signifying the impossibility of a proper landing even with maximum deceleration.

\subsubsection{Action smoothing}
In dealing with high-fidelity plants, it is common for them to account for actuator delay to a certain extent. However, modeling the response perfectly, especially the transient response under oscillating control signals, is impractical. Therefore, for practical actuators, achieving a smooth operating curve is desirable to minimize the disparity between simulation and reality. In this context, RL policies, supported by powerful neural networks, are typically less effective than classical control methods, which is attributed to the high nonlinearity and the absence of intrinsic motivation for smoothness. The challenge becomes more pronounced when addressing goal-oriented tasks, where the requirement of zero intermediate reward hinders direct smoothness regulation.

To address these challenges, two measures are implemented to improve action smoothness in goal-oriented tasks. Firstly, we redefine the action $\Tilde{a}$ as the actuator increment, incorporating the original action $a$ into the state observation $\Tilde{s}$:
\begin{equation}
    \label{eqn:incremental}
    \begin{aligned}
        \Tilde{s} &= \begin{bmatrix}s & a\end{bmatrix} \\
        \Tilde{a} &= \Delta a \\
        a' &= \text{clip}(a + k \Delta a, -1, 1)
    \end{aligned}
\end{equation}
where $k$ is a scaling factor. Secondly, instead of relying on a regulatory reward, we intervene in the learning process at the loss level. This is achieved by adding a term to PPO's policy loss \eqref{eqn:ppo-loss} alongside the advantage:
\begin{equation}
    \label{eqn:aux-loss}
    \Tilde{J}_\pi(\theta) = J_\pi(\theta) + \epsilon \sum \lVert \Tilde{a} \rVert^2,
\end{equation}
where $\epsilon$ is a small positive coefficient. While these two measures significantly reduce oscillation, they also introduce increased learning difficulty. Through cascading jump start, these measures are only applied in the second training stage with a strong guide policy, thereby alleviating the associated difficulties.

\section{Experiment}

\subsection{Environment configuration}
\begin{figure}[h]
    \centering
    \includegraphics[width=0.95\linewidth]{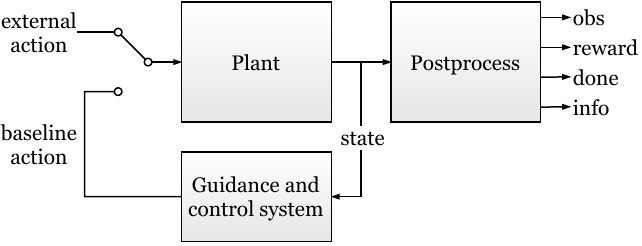}
    \caption{Wrapped plant for RL training}
    \label{fig:schematic}
\end{figure}

As detailed in the problem statement, the high-fidelity rocket plant and its baseline controller were modeled using Simulink. To enable interaction with the RL policy, we wrapped the system to comply with the standard RL environment interface, as is shown in Fig.~\ref{fig:schematic}. RL training demands a significant number of samples for convergence. However, Simulink's interpreted execution is not optimal for efficient and parallel environment interaction. To address this, we utilized Simulink Embedded Coder to generate C code, compiling it into an efficient native module. The use of GOPS Slxpy \cite{wang2023gops} facilitated automated glue code generation, producing a cross-platform binary with deterministic simulation and flexible parameter tunability. Notably, the control signal supplied to the plant can dynamically transition between external action and the baseline controller through a parameter, streamlining integration with RAJS.

In terms of reward formulation, all experiments, except for the trajectory tracking baseline mentioned later, share an intermediate reward of zero and a terminal reward defined as follows:
\begin{equation*}
    r_T = \begin{cases}
        r_\text{prox} & s_T \in \mathcal{S}_{\text{prox}},\\
        0 & s_T \notin \mathcal{S}_{\text{prox}}.
    \end{cases}
\end{equation*}
Here, $r_\text{prox}$ is defined according to \eqref{eqn:log-reward}, with $p = 0.1$ and $b = 3.5$. For the extra early termination condition, $a_{\max,1}$ and $a_{\max,2}$ are set to \qty{40}{m/s^2} and \qty{8}{m/s^2}, respectively, based on preliminary testing with open-loop signals.

\subsection{Benchmark experiment}
In the experiments, we combine RAJS with the on-policy PPO algorithm, referred to as PPO-RAJS, and compare its performance against several baselines, all utilizing PPO as the underlying RL algorithm:
\begin{enumerate}
    \item PPO: This baseline applies the PPO to solve the goal-oriented environment without modification.
    \item PPO-Track: The environment definition is adjusted to track predefined trajectories of the baseline controller while incorporating a shaped tracking reward.
    \item PPO-JSRL: This variant integrates PPO with JSRL-Random, where the baseline controller serves as the guide policy. We excluded JSRL-Curriculum from the benchmark as it is incompatible with on-policy RL.
\end{enumerate}

We provide the hyperparameters of these algorithms in Table \ref{tab:hyperparameters}. To mitigate the influence of randomness, we conduct three experiments for each algorithm using different seeds. The combination of an efficient native environment module and PPO's capability for parallel sampling allows for rapid training, achieving $10^8$ steps in 5 hours.
\begin{table}[t]
    \centering
    \caption{Hyperparameters}
    \label{tab:hyperparameters}
    \begin{tabular}{lll}
        \toprule
        Algorithm & Parameter & Value \\
        \midrule
        \multirow{10}{*}{Shared} & Learning rate & \num{3e-4} \\
        & Network size & $(256,256)$ \\
        & Network activation & tanh \\
        & Discount factor $\gamma$ & \num{0.995} \\
        & GAE $\lambda$ & \num{0.97} \\
        & Train batch size & \num{20000} \\
        & Gradient steps & \num{30} \\
        & Clip parameter $\epsilon$ & 0.2 \\
        & Target KL divergence & 0.01 \\
        & Entropy coefficient & \num{0.007} \\
        \midrule
        \multirow{3}{*}{PPO-RAJS} & Maximum guide horizon $\bar{H}$ & 18 \\
        & Success rate threshold $P_\text{thresh}$ & 0.3 \\
        & Annealing step size $\alpha$ & $1 / 1500$ \\
        \bottomrule
    \end{tabular}
\end{table}

\begin{figure}[t]
    \centering
    \includegraphics[width=0.48\textwidth]{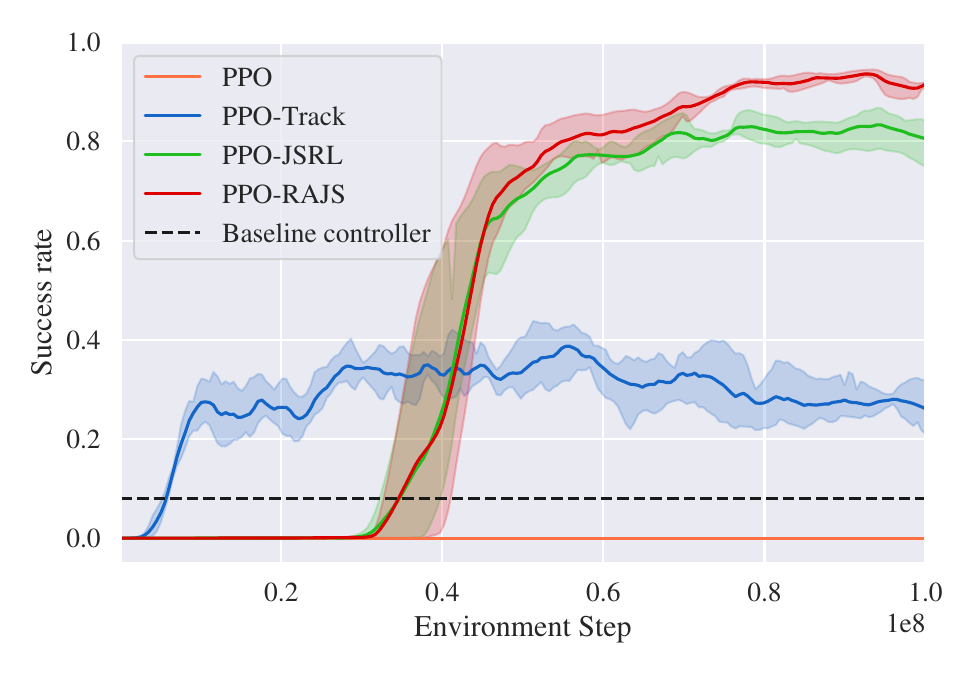}
    \caption{Success rate for PPO-RAJS and baselines. The solid lines correspond to the mean and the shaded regions correspond to 95\% confidence interval over three runs.}
    \label{fig:success-rate}
\end{figure}

We evaluate the performance of these algorithms based on their success rate, defined as the fraction of trajectories that reach the goal set. The learning curves depicted in Fig.~\ref{fig:success-rate} illustrate the following results. Our algorithm PPO-RAJS achieves a high success rate and a small variance across different seeds at convergence. PPO fails to optimize the policy due to the sparse reward signals. PPO-Track has poor performance because of insufficient adaptiveness to different initial condition and disturbance. PPO-JSRL exhibits similar trend compared to PPO-RAJS during the initial rise, but cannot further improve performance for the rest of the run due to distribution mismatch issues.

\begin{figure}
    \centering
    \subfloat[Success rate]{\label{fig:success-rate-2}\includegraphics[width=0.45\textwidth,trim={0.5cm 0.5cm 0.5cm 0.5cm}, clip]{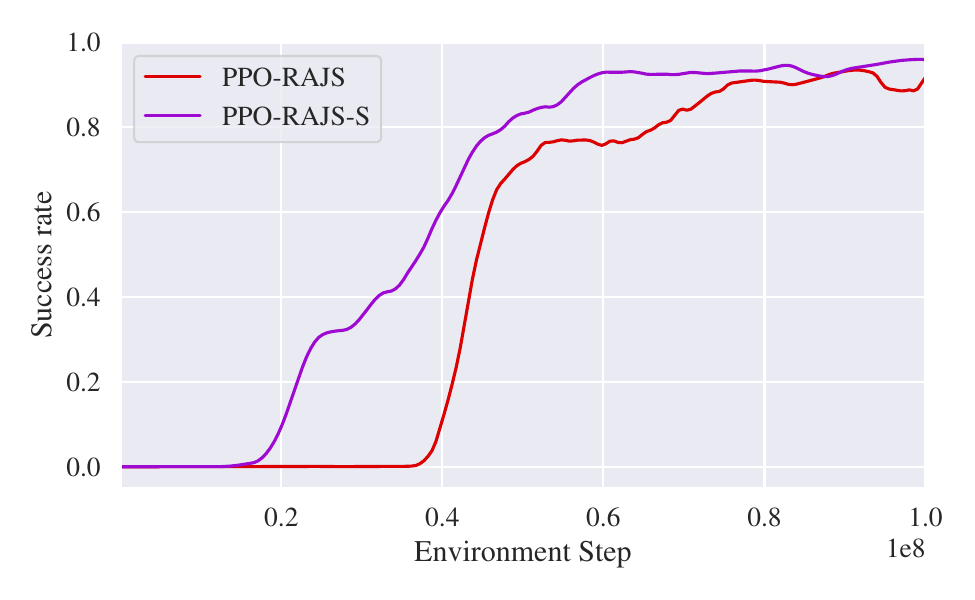}} \\
    \subfloat[Action fluctuation (lower is better)]{\label{fig:fluctuation}\includegraphics[width=0.45\textwidth,trim={0.5cm 0.5cm 0.5cm 0.5cm}, clip]{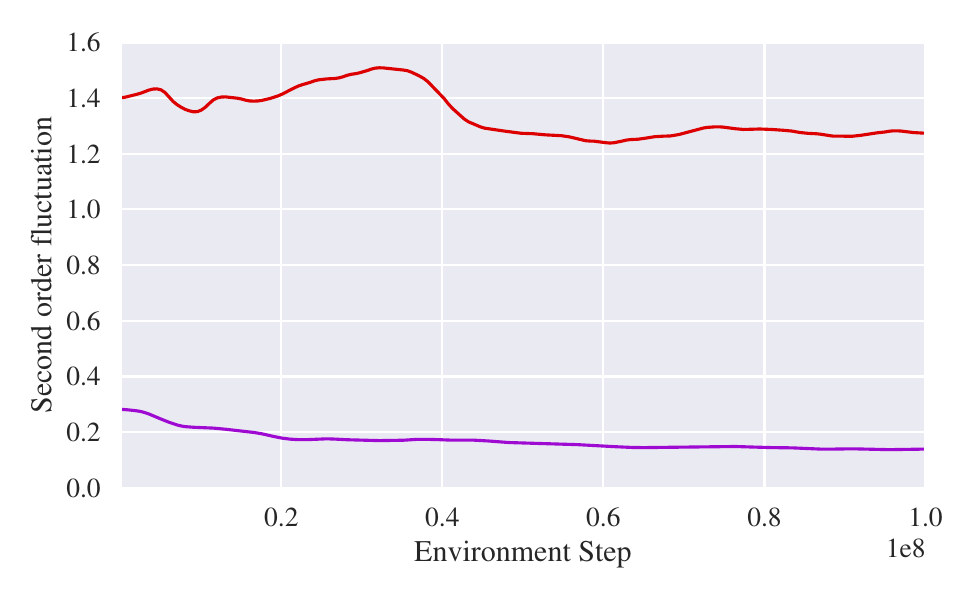}}
    \caption{Training curve comparison between PPO-RAJS and PPO-RAJS-Smooth.}
\end{figure}

The trained policy from PPO-RAJS is then employed as the new guide policy, following the principles of the cascading jump start technique, to learn the smoothness-focused PPO-RAJS-S policy. This policy incorporates equations \eqref{eqn:incremental} and \eqref{eqn:aux-loss} with a smoothness regulation coefficient $\epsilon = 0.01$ and an incremental action scaling factor $k = 0.4$ for all action components. Fig.~\ref{fig:success-rate-2} illustrates that PPO-RAJS-S achieves faster learning and further improves the success rate, even with the additional challenge of action smoothness regulation. To quantify the effect of action smoothness, we introduce the second-order fluctuation $F_2$, defined as:
\begin{equation*}
    F_2 = \frac{1}{T - 2}\sum_{t=2}^{T-1}{|a_t + a_{t-2} - 2a_{t-1}|}.
\end{equation*}
The training curve presented in Fig.~\ref{fig:fluctuation} clearly indicates the advantage of incremental action from the outset, with further improvements in action smoothness observed during training through the simple regulation measure.

\subsection{Evaluation result}
\begin{figure}
    \centering
    \begin{overpic}[width=0.95\linewidth]{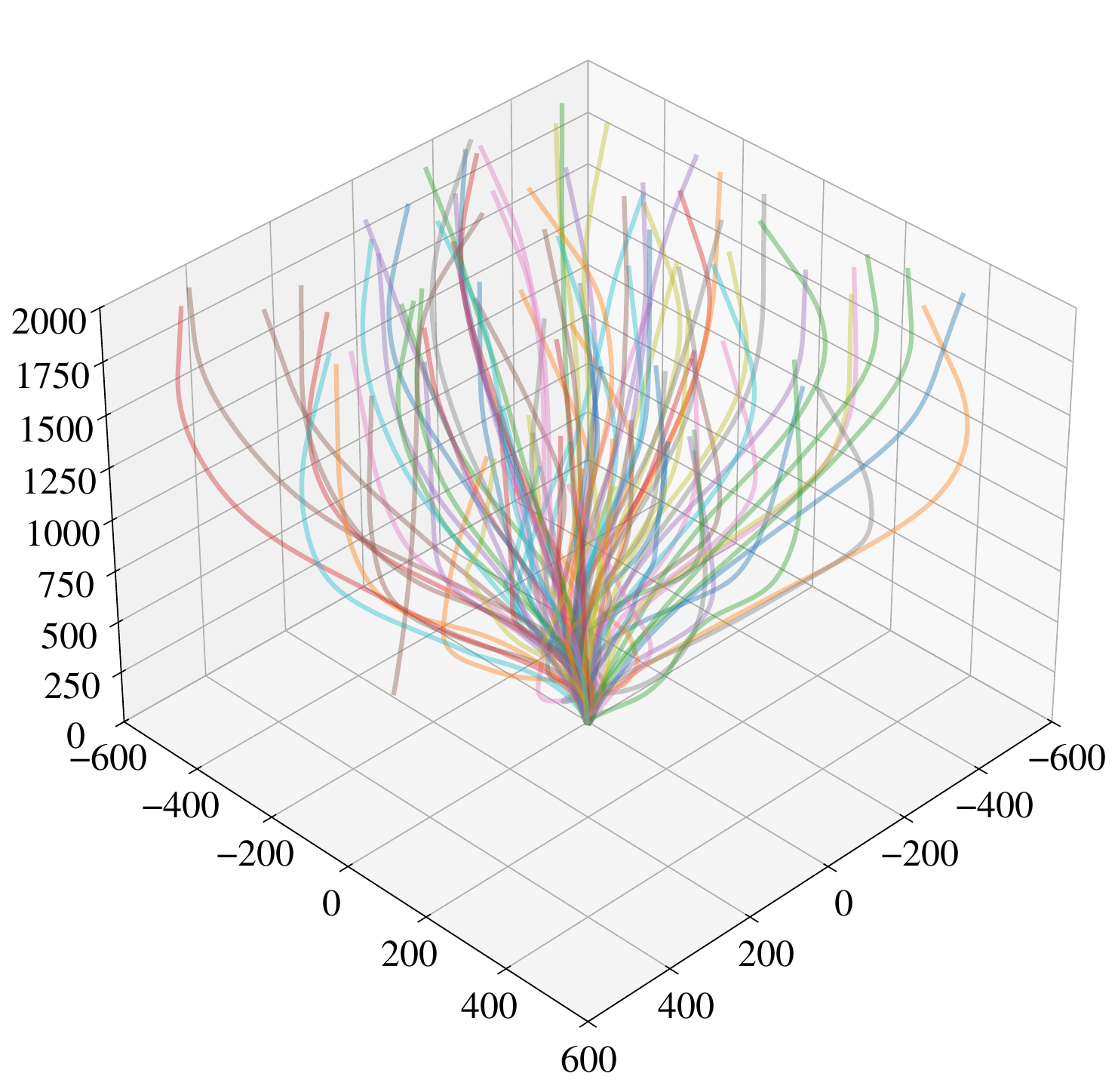}
        \put (-2,45) {\rotatebox{90}{$y$ [m]}}
        \put (18, 13.5) {$z$ [m]}
        \put (77, 13.5) {$x$ [m]}
    \end{overpic}
    \caption{A subset of 100 trajectories controlled by PPO-RAJS-S policy.}
    \label{fig:traj}
\end{figure}

We generate $10^6$ distinct initial conditions to comprehensively evaluate the final performance of the PPO-RAJS-S policy across a vast initial state space. The statistical analysis presented in Table~\ref{tab:statistics} aligns closely with the training curves. Notably, the component $v_y$ exhibits the highest frequency of violations among all constraints, corroborating the earlier discussion on the challenge posed by controlling $v_y$ effectively. Figure~\ref{fig:traj} illustrates that the majority of trajectories either achieve success or fail proximal to the goal set boundary. However, a small subset experiences pose instability due to aggressive pose control, leading to landing significantly distant from the target, highlighting a current limitation of our approach.
\begin{table}
    \centering
    \caption{Final policy evaluation statistics}
    \label{tab:statistics}
    \begin{tabular}{ccc}
        \toprule
        Success rate & \multicolumn{2}{c}{0.9739} \\
        Landing rate & \multicolumn{2}{c}{0.9953} \\
        \midrule
        \multicolumn{3}{c}{(Within non-landing trials)} \\
        Vertical speed reversal & \multicolumn{2}{c}{0.0043} \\
        Fuel exhaustion & \multicolumn{2}{c}{0.0003} \\
        \midrule
        (Within landing trials) & Satisfaction rate & Error 99\textsuperscript{th} percentile \\
        $x$        & 0.9930 & 4.5849 \\
        $z$        & 0.9944 & 4.2008 \\
        $v_x$      & 0.9934 & 0.8599 \\
        $v_y$      & 0.9831 & 1.0869 \\
        $v_z$      & 0.9939 & 0.8338 \\
        $\phi$     & 0.9949 & 1.9780 \\
        $\psi$     & 0.9957 & 1.3992 \\
        $\dot\phi$ & 0.9955 & 0.6678 \\
        $\dot\psi$ & 0.9955 & 0.6511 \\
        \bottomrule
    \end{tabular}
\end{table}

\begin{figure}
    \centering
    \includegraphics[width=0.9\linewidth]{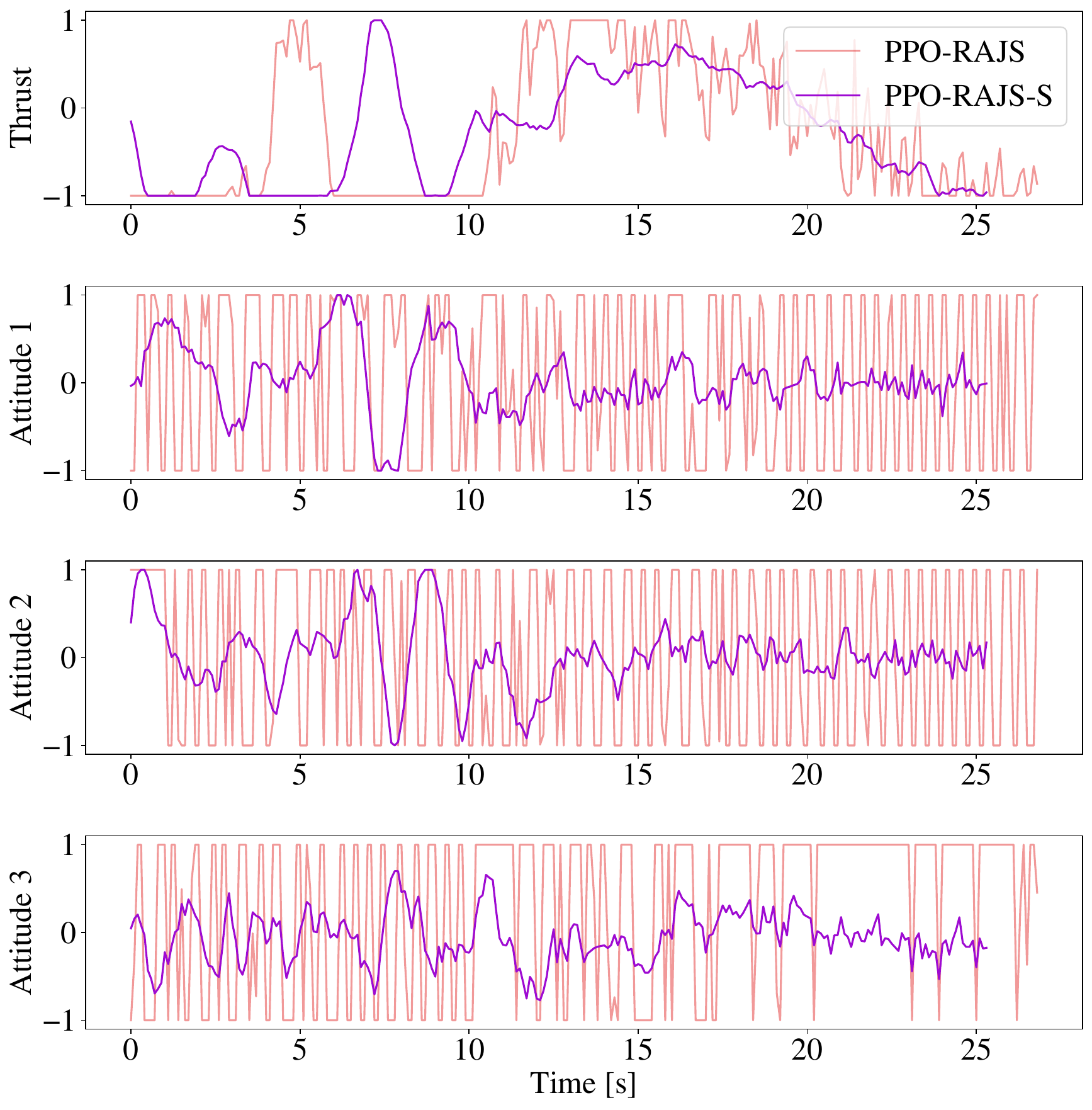}
    \caption{Action sequence comparison between PPO-RAJS and PPO-RAJS-S, starting from a common initial state.}
    \label{fig:smooth}
\end{figure}

The enhanced smoothness of actions is depicted in Figure~\ref{fig:smooth}. While both PPO-RAJS and PPO-RAJS-S successfully reach the goal set, the former displays notable fluctuations, which could potentially challenge practical actuator implementations. Conversely, actions produced by PPO-RAJS-S exhibit considerably smoother behavior.

Furthermore, we deploy the model on the ZU9E embedded platform and perform hardware-in-loop co-simulation with a rocket dynamics simulation engine. Results from the simulation demonstrate that policy inference aligns within the control interval of \qty{10}{ms}, validating its real-time applicability.

\section{Conclusions}
This paper presents the random annealing jump start approach, which utilizes baseline controllers to empower RL algorithms in tackling complex real-world goal-oriented tasks. Given the safety-critical nature of rocket landing control, our future research will delve into integrating safe RL theory, such as neural barrier certificate \cite{yang2023model},  to manage state constraints more effectively. This integration holds promise in addressing the current challenge of unguaranteed pose stability and in further augmenting task success rates.

\section*{Acknowledgment}
We extend our gratitude to our colleagues at LandSpace for providing their rocket model, contributing innovative insights, and offering consistent support throughout the entirety of this research endeavor, without which this study would not have been possible. LandSpace Technology Co.,~Ltd.\ (LandSpace) was founded in 2015 and is a leading Chinese enterprise in the creation and operation of space transportation systems. It is dedicated to establishing a comprehensive industrial chain encompassing ``research and development, manufacturing, testing, and launching,'' with a focus on medium and large Liquid Oxygen-Methane launch vehicles. LandSpace aims to create a technological hub in the space sector and provide highly cost-effective and reliable space transportation services to the global market.

\bibliography{refs}

\end{document}